\documentclass{article} %
\usepackage{iclr2022_conference,times}

\usepackage{amsmath,amsfonts,bm}

\def\eqref#1{equation~\ref{#1}}

\def\1{\bm{1}}

\DeclareMathAlphabet{\mathsfit}{\encodingdefault}{\sfdefault}{m}{sl}
\SetMathAlphabet{\mathsfit}{bold}{\encodingdefault}{\sfdefault}{bx}{n}

\newcommand{\E}{\mathbb{E}}

\newcommand{\KL}{D_{\mathrm{KL}}}

\usepackage{hyperref}
\usepackage{url}
\usepackage[utf8]{inputenc} %
\usepackage[T1]{fontenc}    %
\usepackage{booktabs}
\usepackage{tabularx}
\usepackage{diagbox}       %
\usepackage{subfig} %
\usepackage{multirow}

\usepackage[disable]{todonotes}
\newcommand{\addcite}[1]{\todo[color=green!40]{\tiny Ref: #1\par}}

\definecolor{CMlightpurple}{rgb}{0.87,0.72,0.88}
\definecolor{CMpurple}{rgb}{0.6,0.18,0.64}

\newcommand\cms{\bgroup\markoverwith{\textcolor{CMpurple}{\rule[.4ex]{2pt}{0.8pt}}}\ULon}

\definecolor{CPblue}{HTML}{03A9F4}

\title{\textsc{Hindsight}: Posterior-guided training of retrievers for improved open-ended generation}

\preprintcopy %

\author{Ashwin Paranjape, Omar Khattab,\\ \textbf{Christopher Potts, Matei Zaharia \& Christopher D. Manning} \\
Stanford University\\
\texttt{\{ashwinp,okhattab\}@cs.stanford.edu} }

\def\zg{{z_{gold}}}
\def\Pgen{{P_\theta(y|x,z)}}
\def\Pret{{P_\eta(z|x)}}
\def\Q{{Q(z|x,y)}}
\begin{document}
\maketitle
\begin{abstract}
Many text generation systems benefit from using a retriever to retrieve passages from a textual knowledge corpus (e.g., Wikipedia) and providing these passages as additional context to the generator. For open-ended generation tasks (like generating informative utterances in conversations) many varied passages may be equally relevant and we find that existing methods that jointly train the retriever and generator underperform: the retriever may not find relevant passages even amongst the top-10 and the generator may hence not learn a preference to ground its generated output in them. We propose using an additional guide retriever that is allowed to use the target output and ``in hindsight’’ retrieve relevant passages during training. We model the guide retriever after the posterior distribution Q of passages given the input and the target output and train it jointly with the standard retriever and the generator by maximizing the evidence lower bound (ELBo) in expectation over Q.  For informative conversations from the Wizard of Wikipedia dataset, with posterior-guided training, the retriever finds passages with higher relevance in the top-10 (23\% relative improvement), the generator’s responses are more grounded in the retrieved passage (19\% relative improvement) and the end-to-end system produces better overall output (6.4\% relative improvement).

\end{abstract}

\section{Introduction}
\label{sec:introduction}

In knowledge-intensive NLP tasks, models must use open-domain knowledge to answer questions \citep{NQ, trivia-qa}, fact-check claims \citep{FEVER}, engage in informative conversations \citep{WoW}, or otherwise facilitate information access through natural language \citep{MSMarco}.
To leverage the information present in human-readable corpora (e.g., Wikipedia), many recent models for open-domain question answering \citep{ORQA,colbert-QA} are \textit{retrieval-augmented}: they extract passages from the corpus using a learned \textit{retriever} and process it with a task-specific \textit{reader}. 
If the relevant passage is known, the retriever is supervised to find it; otherwise, it is weakly supervised with passages that contain the answer string (e.g., a name or a date).
In recent work \citep[e.g.,][]{RAG}, for end-to-end tasks with arbitrary text output, where relevant passages are not known, a jointly-trained generative sequence-to-sequence model provides the relevance signal.

\begin{figure}
  \centering
    \includegraphics[width=0.8\textwidth]{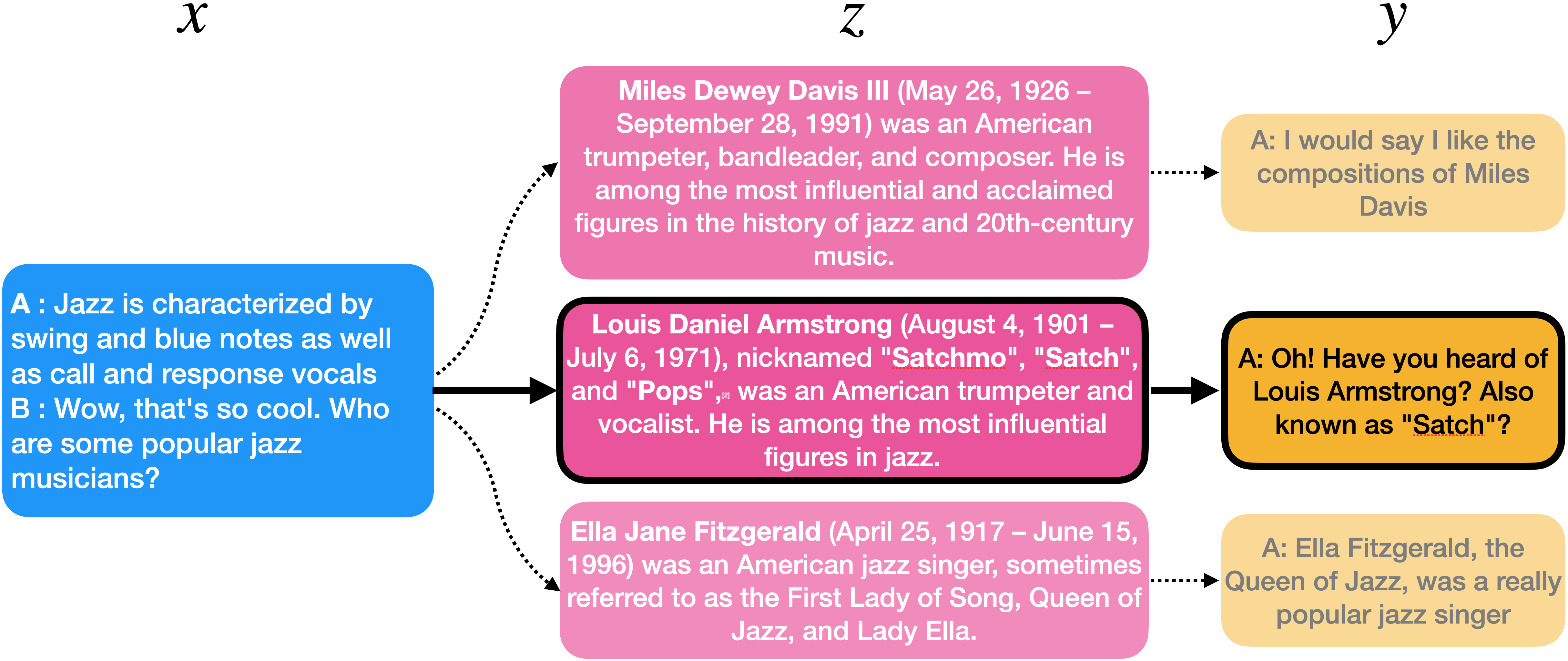}

    \caption{A conversational turn with multiple plausible responses. The input (blue) can be answered based on 3 equally relevant passages but only one possible response (yellow) is observed in the training set (shown outlined in black) based on only one of the pink relevant passages (black outline). }
    \label{fig:setting_example}
\end{figure}

Current methods work well for short-answer QA-like problems, where the query itself has high overlap with the relevant passage and there is a unique answer, for example factoid Open-domain QA such as Natural Questions \citep{NQ}, HotPotQA \citep{hotpotQA}, fact-checking \citep{FEVER}, and slot-filling \citep{ZeroShotRE}. 
However, for \textit{open-ended} generation tasks like free-form QA (MS-Marco NLGen), informative dialogue \citep{WoW}, or Wikipedia abstract generation \citep{WikiSum},  
many responses can be equally acceptable and yet the data source (e.g., human-written answers, real-world conversations, Wikipedia abstracts) supplies only one plausible output from a space that includes many.  
Figure~\ref{fig:setting_example} illustrates this with a conversational context $x$ and multiple relevant passages $z$: while only one of them $\zg$ is useful for generating the gold-standard answer $y$, other passages could have led to other coherent responses. 
This setting with one input and many possible unobserved outputs (\textbf{one-to-many}) is challenging, as it creates a distinction between the numerous passages that are relevant to the input (which we dub \textbf{context-relevant} passages) and the few passages that pertain to the observed target output (which we dub \textbf{label-relevant} passages).
Had we known $\zg$ corresponding to the target output, we could have supervised the retriever with $\zg$ and trained the generator conditioned on $\zg$, but we don't!

Current methods \citep{RAG} attempt to use the generator's probability distribution $\Pgen$ as a proxy for label relevance and train the retriever $\Pret$ by marginalizing $p(y|x)$ over retrieved documents $z$: $P(y|x) = \sum_{z \in \text{top-k}(P_\eta(.|x))} P_\eta(z| x) P_\theta(y|x, z)$.
We find empirically that for one-to-many tasks, this objective is suboptimal in three ways: the generator is less grounded in the retrieved passages (Figure~\ref{fig:relevance_groundedness_comparison}), the retriever performance saturates at low recall (Figure~\ref{fig:relevance_groundedness_comparison}), and at training time the top-k retrieved passages miss out on many label-relevant passages weakening the supervision (Table~\ref{table:relevance_evaluation}).

We propose explicitly estimating the label-posterior distribution $\Q$ in the form of a \textbf{guide-retriever} that has access to the target output, and with this ``hindsight'' can capture label-relevance precisely.
We jointly optimize the retriever, posterior-guide, and generator using the evidence lower bound (ELBo): $\E_{z_i\sim Q(.|x, y)} [\log P_\theta(y|x, z)] - \KL (Q \vert P_\eta)$.
In theory, posterior-guided training is better for joint-training on open-ended tasks on all three counts: (1) The generator gets conditioned on passages weighted directly by their label-relevance from the label-posterior distribution, reducing the tendency of the generator to ignore the retrieved passage; 
(2) The retriever is trained with reverse-KL divergence, whose mode-seeking nature encourages the retriever's probability distribution to match some modes with the guide (label-relevant passages), without explicitly penalizing other modes (other context-relevant passages);
(3) As the label-posterior distribution $\Q$ is modeled by a full-fledged retriever, it can retrieve label-relevant passages from the entire collection, which is a generalization of weak supervision approaches that use span overlap for short-answer QA.

Our main contribution in this paper is the \textsc{Hindsight} training system that: (1) uses a guide-retriever to provide a stronger learning signal for both the generator and the retriever, (2) is amenable to index-updates with iterative closed-set training (Section~\ref{sec:hindsight_method}). 
We thoroughly evaluate the retrieval-augmented systems, going beyond end-to-end performance with a granular evaluation of the individual models (retriever and generator) at varying passage depths, important for evaluating one-to-many open-ended generation tasks (Section~\ref{sec:exp_eval}). 
Using \textsc{Hindsight} on the Wizard of Wikipedia dataset of informative conversations our retriever achieves a 23\% relative improvement in success@10 (its ability to have the label-relevant passage among the top-10 retrieved passages), our generator is more grounded with 19\% improvement in Novel-F1 overlap with the top-1 retrieved passage  (i.e., its overlap with rare words from the retrieved passage that were not already in the input) and 6.4\% relative improvement in Novel-F1@1 overlap with the gold utterance (the best matching generation when considering top-1 retrieved passage).
We also validate that \textsc{Hindsight} improves performance on the one-to-one generation task of free-form QA using the MS-MARCO NLGen dataset.

\section{Background}

\paragraph{Open-domain Question Answering}
While Question-Answering datasets originally focused on extracting the answer from a given passage (a.k.a.\ reading comprehension), recent work has made the task more challenging by not supplying a gold passage but instead expecting the model to conduct open-domain QA directly over a large document collection.
The first neural system for answering open-domain factoid questions was Dr.~QA \citep{DRQA}, which used an off-the-shelf retriever (e.g., TF-IDF, BM25) to retrieve relevant passages and a separately trained reader to extract the answer span.  
The advent of efficient nearest-neighbour search algorithms \citep{IVFADC,FAISS} opened up the possibility of encoding the query and the passage collection into a vector space and finding the relevant passage as nearest neighbors to the query embedding.
Subsequent work trained neural retrievers in various ways such as: pretraining with the inverse cloze task then weakly supervising using span matches \citep{ORQA}, using gold passages with in-batch negatives \citep{DPR}, and retrieval-guided supervision with span-based positives \citep{colbert-QA}. 

\paragraph{Open-ended Generation}
Natural language generation tasks involve generating a sequence of tokens (maybe, word-pieces), often contextualized on some input (another sequence of tokens for sequence-to-sequence tasks, an image for an image captioning task).
However, they differ in how open-ended they can be.
Some tasks such as factoid question-answering having a single correct short answer are less open-ended than free-form long answers. 
For example, in informative dialogue the speakers can lead the conversation in many different directions \citep{WoW}, also referred to as one-to-many generation. 
Thus it is more open ended, or, equivalently, it has a higher entropy than say machine translation \cite{WMT2014} which has few correct translations that are very similar to each other. 
Many more generation tasks such as summarization \citep{XSUM} and story generation \citep{rocstories} lie on this spectrum. 

\paragraph{Retrieval for Language Modeling}
To improve the perplexity of a pre-trained language model, \cite{KNNLM} retrieve similar contexts from the training set at each time-step and increase the likelihood of tokens that were predicted in similar contexts.  
\cite{REALM} instead pre-train a retrieval-augmented masked language model using salient-span masking and fine-tune it on downstream QA tasks. 
Under the paradigm of retrieval-augmented generation, for input $x$ and output $y$, \cite{RAG} retrieve top-k passages ($z$) from a corpus with a retriever and jointly train the generator ($P_\theta$) and the retriever ($P_\eta$) by maximizing the likelihood of the output marginalized over the top-k documents, which we refer to as the \textsc{MarginalizedLoss}:
\begin{equation} 
    \label{eq:margloss}
    P(y|x) = \sum_{z \in \text{top-k}(P_\eta(.|x))} P_\eta(z| x) P_\theta(y|x, z)
\end{equation}
Here $\Pgen$ has two roles: supervision, i.e. providing label-relevance by scoring label-relevant passages higher than other passages and grounding, i.e. maximizing the probability of the target output given context-relevant passages. 
For one-to-many datasets, where few context-relevant passages are label-relevant, the two roles are at odds with each other. 
By increasing label-relevance signal the generator reduces the ability to ground in passages that are context-relevant but not label-relevant. 
By increasing grounding in all context-relevant passages it reduces the label-relevance signal. 
In the next section, we describe our method we separates these two concerns, leading to improved performance.

\section{Training with Hindsight}
\label{sec:hindsight_method}
To precisely identify label-relevant passages, we propose explicitly modeling the posterior distribution: $\Q$ with a learned neural model. 
Unlike the retriever, the label-posterior model has access to the target output and in hindsight can differentiate label-relevant from other context-relevant passages. 
We learn the label-posterior jointly with the retriever and the generator by maximizing the evidence lower bound, \textsc{ELBoLoss}, as given by the formula:%
\begin{equation} 
    \label{eq:elboloss}
    \log P(y|x) \ge \E_{z\sim Q(.|x, y)} [\log P_\theta(y|x, z)] - \KL (Q \vert P_\eta)
\end{equation}

For an intuitive understanding we look at the two terms separately. The first term is an expectation of the generator's log-likelihood $P_\theta$ over the label-posterior $Q$. This ensures that the generator need attend only to the label-relevant passages and therefore learns to ``trust'' the retrieved passages better. 
The second term is the KL divergence from the retriever to the label-posterior, also referred to as reverse KL divergence:
\[\KL\big[\Q \mid \Pret\big] = \sum_{z\sim Q(.|x, y)}\Q \bigl(\log \Q -  \Pret\bigr)\]
This term is again weighted by $\Q$, which is the probabilistic equivalent of implication: high $Q(z|x,y)$ implies high $P(z|x)$, i.e., label-relevance implies context-relevance but not vice-versa.

\paragraph{Posterior as a retriever}
While we can certainly model the label-posterior $Q(z|x,y)$ as a re-scorer that takes in documents as retrieved by the retriever $P_\eta$, we instead model it as a guide retriever that can retrieve label-relevant passages from the entire corpus.
This is helpful because we can sample passages directly from the label-posterior distribution, and estimate the \textsc{ELBoLoss} accurately than using passages from $\Pret$. 
The guide retriever generalizes weak supervision approaches \citep{ORQA,REALM} and relevance-guided supervision \citep{colbert-QA}, to \textbf{posterior-guided supervision} with a learned posterior retriever rather than brittle heuristics based on word-overlap.

\begin{figure}
  \centering
    \includegraphics[width=\textwidth]{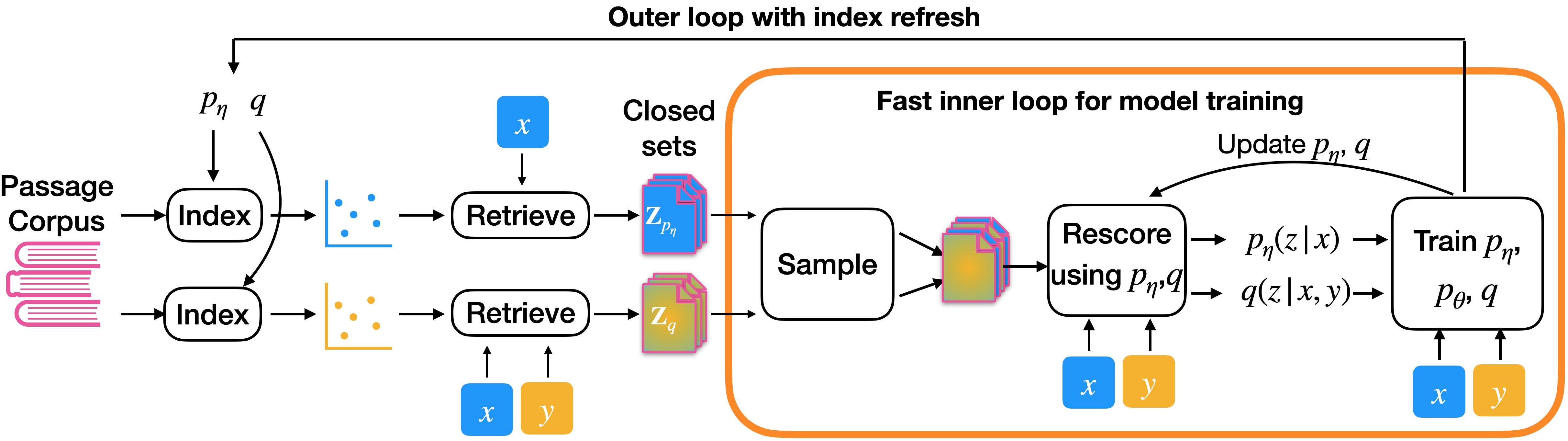}
    \caption{An overview of iterative closed-set training: We iterate through the outer-loop and call each execution a round. At the beginning of the round we re-index the passage corpus using the latest retriever $p_\eta$ and guide-retriever $q$ to create a high-recall closed-set of top-$r$ passages for each retriever and query. Then, in the fast inner loop, we train the models for multiple epochs by sampling passages from the fixed closed-set and recomputing the probability distributions. The trained models are then used in the next round.
    }
    \label{fig:inner_outer_loop}
\end{figure}

\paragraph{Iterative closed-set training}
Prior work \citep{REALM, colbert-QA} has shown the utility of intermittently updating the passage index during training.
To allow for such a workflow, we organize our training into rounds (see Figure~\ref{fig:inner_outer_loop}). 
At the beginning of each round, in the outer loop we encode the passages and the queries with various retrievers and find the highest scoring $r$ passages that we dub the closed-set. 
In the inner loop that runs for many epochs, we sample passages from the closed-set ($r=100$ for our experiments). 
This is fast because we are no longer retrieving from the entire corpus in the inner loop and also sufficient because the closed-set has a high recall. 
We update the retrievers (both document and query encoders) during the inner loop and use the latest model parameters for computing the loss functions. 
A round results in trained models that are then used for the next round.
We find that 2 rounds are often sufficient, with decreasing marginal utility from the third round onward.

\paragraph{Distributional repositioning before inference}
According to \textsc{ELBoLoss}, the expectations are computed over $z \sim Q(.|x,y)$.
We cannot compute the expectations exactly because of the large size of the passage corpus, so we approximate it by sampling $k$ passages from the closed-set $Q_{\text{top-r}}(Q(.|x,y))$. 
This leads to faster model training, especially at the beginning because passages from $Q(.|x,y)$ provide better supervision. 
However, due to this sampling scheme, the models only ever get exposed to passages from the $Q(.|x,y)$ distribution, which limits their ability to generalize over passages from $P_\eta(.|x)$ during inference. 
To remedy this, we sample passages from an $\alpha$-mixture  of the two distributions: $z \sim \begin{cases} 
      P_\eta(.|x) & \text{with prob.} ~\alpha \\
      Q(.|x,y) & \text{with prob.} ~1-\alpha 
   \end{cases}$.
In the initial rounds we set low values of $\alpha$ and increase it toward the end to reposition the passage distribution and better match with $P_\eta(.|x)$ at test time. 

\paragraph{Training individual models to convergence}
A practical issue with joint training is that the retriever and generator train at different rates, reaching convergence at different times.
The single term in \textsc{MarginalizedLoss} (Eq.~\ref{eq:margloss}) does not indicate convergence of individual models leading to situations where one model starts to overfit while the other model still hasn't converged. 
With \textsc{ELBoLoss} there are two terms in Eq.~\ref{eq:elboloss}: the first connecting $\Q$ and $\Pgen$, the second connecting $\Q$ and $\Pret$. 
This allows us to fix the guide and train the retriever and generator independently for a differing number of gradient steps until each model reaches convergence.

\section{Experimental Evaluation}
\label{sec:exp_eval}
\begin{figure}
  \centering
    \subfloat{{\includegraphics[width=2.5in]{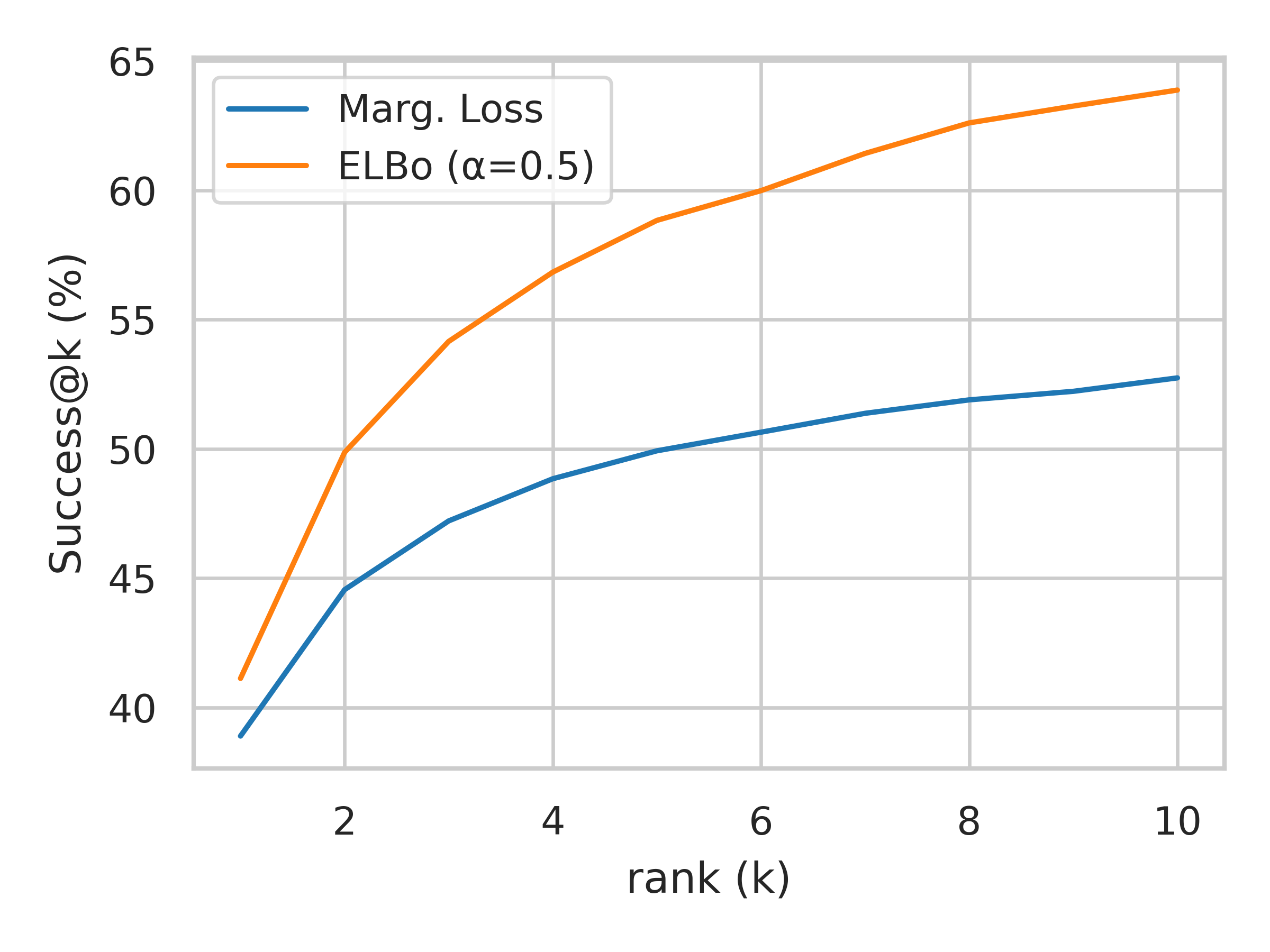}}}
    \qquad
    \subfloat{{\includegraphics[width=2.5in]{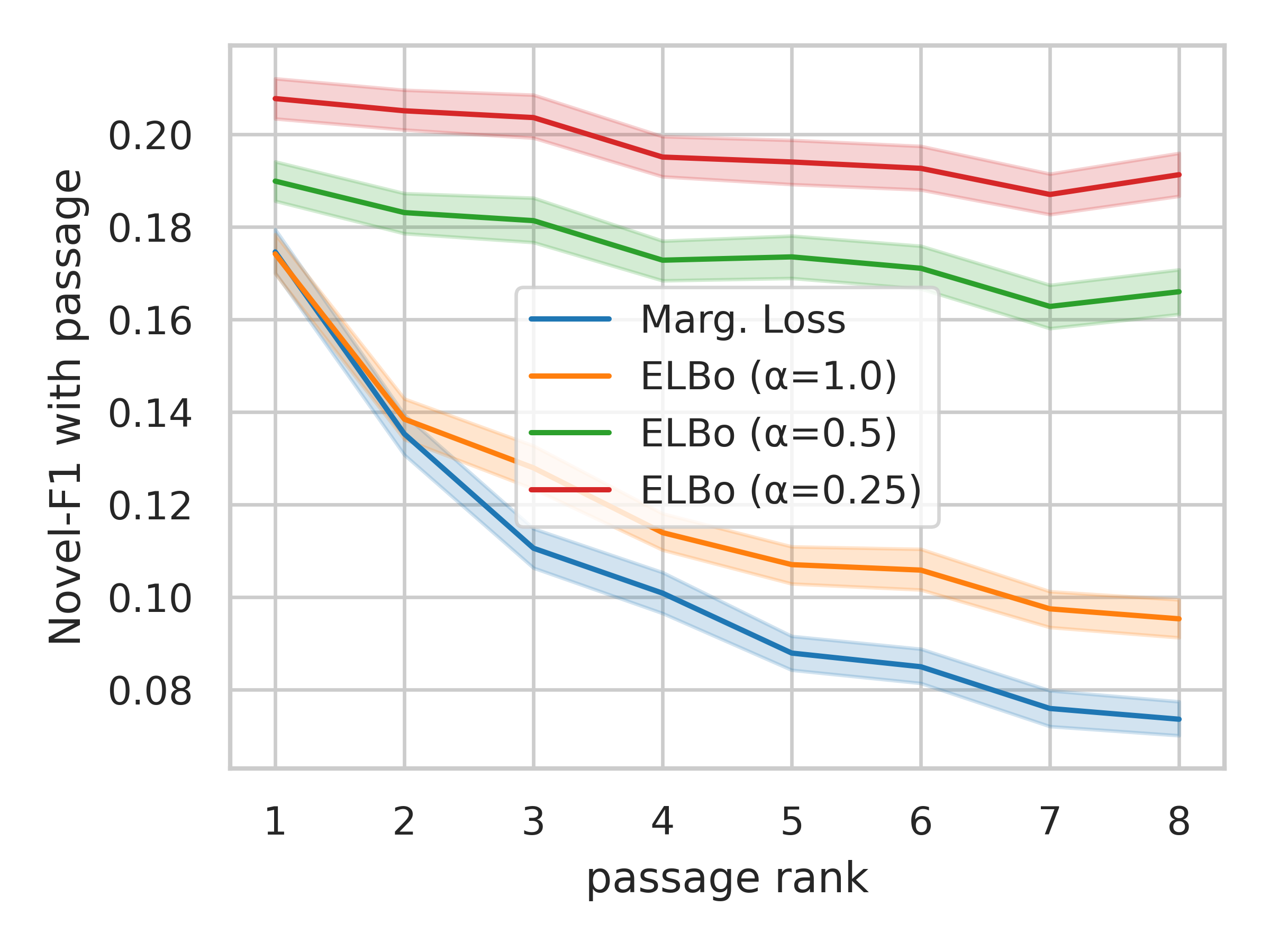}}}
    \caption[]{Relevance and Groundedness of models trained on the Wizard of Wikipedia dataset: (\textbf{left}) success@k of retrieved passages w.r.t. rank and (\textbf{right}) Novel-F1 between decoded output and retrieved passage w.r.t retrieved passage rank. The \textsc{ELBoLoss} retriever is more effective at retrieving the gold passage than \textsc{MarginalizedLoss} retriever, especially when we consider the top-10 passages for this one-to-many task. The \textsc{ELBoLoss} generators have higher overlap with top-k retrieved passages and the overlap increases as $\alpha$ decreases}
    \label{fig:relevance_groundedness_comparison}
\end{figure}
We evaluate on two open-ended knowledge-intensive tasks: informative conversations and free-form question answering. We ask the following three research questions:
\begin{description}\setlength{\itemsep}{0pt} 
\item[RQ$_1$] Relevance: Are the retrieved passages more relevant? (Section \ref{sec:relevance_evaluation})
\item[RQ$_2$] Groundedness: Does the generator make better use of the retrieved passages? (Section \ref{sec:groundedness_evaluation})
\item[RQ$_3$] Generation Quality: Does this lead to better end-to-end performance? (Section \ref{sec:generation_quality_evaluation})
\end{description}

\subsection{Models}

\paragraph{Retriever Models} We use ColBERT \citep{colbert} to model the retriever $\Pret$ and the guide-retriever $Q$. 
The query tokens $q_i$ and the document tokens $d_j$ are independently encoded using BERT and normalized to produce unit-vectors $E_{q_i}$ and $E_{d_j}$. 
The similarity between the query and the document is defined as $S_{q,d} = \sum_{i} \max_{j} E_{q_i} E_{d_j}^T$. 
The late-interaction paradigm of ColBERT is more expressive than using just the [CLS] token, as in DPR \citep{DPR}, because it allows the query and document tokens to retain their identities and provide finer-grained term-wise similarity. Recently, \citet{colbert-QA} has shown that ColBERT establishes state of the art retrieval results on open-domain QA benchmarks.
To calculate the probability distributions during training, we sample a small set of documents $\mathcal{R}$ using a suitable sampling strategy (as explained in Section~\ref{sec:hindsight_method}) and compute the softmax of the scores over $\mathcal{R}$.
For the posterior-retriever, we concatenate the input and the output to create the query, $q=[x \ y]$.
For the Wizard of Wikipedia task, we use a ColBERT model that has been pre-trained on the MS-MARCO passage ranking dataset. 
However, for the MS-MARCO NLGen task, we use another ColBERT model that has been pre-trained using the Natural Questions dataset instead. 

\paragraph{Generation Model} Following \cite{RAG} we use pre-trained BART model and fine-tune it for the respective tasks during training. 
It is conditioned on both the context and the document and trained to produce the target. At test time, we decode using beam-search with 4 beams.

\subsection{Tasks}
\paragraph{Informative conversations} 
We first evaluate on generating utterances in informative conversations as a one-to-many generation task that is open-ended and knowledge-intensive. 
Informative conversations are one-to-many because they are open-ended and people have the agency to drive the conversation in a different direction at every turn.
We use the Wizard of Wikipedia (WoW) dataset \citep{WoW}, where an ``apprentice'' chats (via text) with a ``wizard'', being curious about different topics, and the ``wizard'' grounds their response in a sentence from Wikipedia. 
The input for this task is the conversational history $x$ and the output is the wizard's utterance $y$ with the models provided access to individual passages ($z$) from all of Wikipedia ($\approx$26 million passages). 
We use the version of this dataset provided in the KILT benchmark \citep{kilt}. 
As the test set outputs are not available to us, we show granular evaluation numbers on the dev set and report leaderboard performance on the test set. 
\addcite{Need to say that RAG evaluated on it, but didn't look at relevance judgements and only looked at overlap with answers}

\paragraph{Free-form Question Answering}
We also validate the efficacy of our method on on-to-one generation task: free-form open-domain QA.
We use the MS-MARCO NLGen dataset \citep{MSMarco} where the task is to generate natural-sounding answers to questions, which is more challenging than other \emph{extractive} open-domain QA datasets. 
This is a subset of MS-MARCO questions whose answers were reviewed by a separate editor and rewritten if there was a high overlap between the answer and one of the provided passages (indicating that the original editor may have copied the passage directly). 
These ``well-formed answers'' are meant to be complete sentences (such as can be read out by a conversational assistant) and have a median length of 11 words. 
The input for this task is a query $x$, the output is a well-formed answer $y$, and the models are required to retrieve from the entire collection provided with the MS-MARCO QA consisting of 8.8 million passages from the web. As the public benchmark is no longer available we could not evaluate on the benchmark test set. Instead we split the public validation set into a validation and test set and show results on the test set. %

While both datasets annotate the passages referred to by the person who wrote the target output (gold passages), we only use them for evaluation and validation and not for training nor for validation.

\subsection{Relevance Evaluation}
\label{sec:relevance_evaluation}

\begin{table}
  \caption{Relevance evaluation: Our method (\textsc{ELBoLoss} Retriever, $\alpha=1$) strongly improves over the baseline (\textsc{MarginalizedLoss} Retriever) for the one-to-many Wizard of Wikipedia dataset, in particular for $k=5, 10$. Our method shows smaller but consistent improvements on the one-to-one MS MARCO NLGen dataset. The ELBo posterior finds $\zg$ with high success providing better supervision during training. (MRR = Mean Reciprocal Rank, Success@k both in percentages)}
  \label{table:relevance_evaluation}
  \centering
  \begin{tabular}{l@{\hspace*{2em}}rrrr@{\hspace*{3em}}rrrr}

    \toprule
    &\multicolumn{4}{c@{\hspace*{3em}}}{Wizard of Wikipedia} & \multicolumn{4}{c}{MS MARCO NLGen} \\
    Method  &   MRR &  S@1    & S@5 & S@10 & MRR & S@1 & S@5 & S@10 \\
    \midrule
    Marg. Retriever & 43.8 & 38.9 & 49.9 & 52.8 & 30.4 & 19.4 & 43.4 & 53.2 \\
    ELBo Retriever   & \textbf{49.0}  & \textbf{41.1} & 
    \textbf{58.8} & \textbf{63.9} & \textbf{32.1} & \textbf{21.2} & \textbf{45.3} & \textbf{54.4}\\
    \midrule
    ELBo Posterior & 78.5 & 72.4 & 86.0 & 88.4 & 67.8 & 56.7 & 81.9 & 86.2\\

    \bottomrule
  \end{tabular}
\end{table}

We begin by investigating the quality of the retrieved passages (\textbf{RQ$_1$}), leveraging the gold passage labels supplied by each dataset. To evaluate relevance, we report Success@$k$ (S@$k$ for short), the percentage of inputs for which the gold provenance passage is retrieved within the the top-$k$ passages by each system. While multiple passages may lead to the desired output, our datasets mark only a single passage as the gold provenance for each input, and so we report success at retrieval depths $k=\{1,5,10\}$.
We also report Mean Reciprocal Rank (MRR), an evaluation metric commonly used to evaluate IR systems.

Our results are shown in Table~\ref{table:relevance_evaluation}. Starting off with Wizard of Wikipedia, we observe that our \textsc{ELBoLoss} retriever markedly outperforms \textsc{MarginalizedLoss} at finding the label-relevant passage. While both systems easily handle 38.9--41.1\% of the examples (evident in relatively high success@1), only the \textsc{ELBoLoss} retriever continues to find many more relevant passages at larger retrieval depths $k$, reaching success@10 of 63.9\%. This considerably exceeds \textsc{MarginalizedLoss}'s success@10 (at 52.8\%). Interestingly, the success@5 rate of \textsc{ELBoLoss} is, in fact, higher than \textsc{MarginalizedLoss}'s success@100 (not reported in the table), where the latter saturates at 55.8\% despite containing 20$\times$ more passages. In contrast, \textsc{ELBoLoss}'s success@100 reaches 69.3\%.

To explain the much stronger success@$k$ scores for $k > 1$ under \textsc{ELBoLoss}, we hypothesize that besides straightforward examples where both systems find a unique ``best'' passage, the label-relevant passages are often less apparent and, in those cases, the \textsc{ELBoLoss} retriever's better-refined training signal allows us to find provenance for more challenging examples. This is confirmed by our informal inspection of results, where we found that the sets of passages retrieved by the two methods on Wizard of Wikipedia are qualitatively different. While not captured by word-overlap measures, we found that the \textsc{MarginalizedLoss} retriever would select many similar ``safe'' passages while the \textsc{ELBoLoss} retriever would select a more diverse and rewarding set of passages (see examples in Appendix~\ref{sec:retrieved_passage_examples}).

Shifting our attention to MS MARCO NLGen, we notice that \text{ELBoLoss} still outperforms \text{MarginalizedLoss} by 1--2 points across our metrics, reflecting smaller---but nonetheless consistent---gains when compared with the one-to-many generation task. We also observe that success@1 is lower for both methods, when compared with Wizard of Wikipedia, a fact we largely attribute---based on manual inspection---to the presence of many false negatives, i.e., passages that contain the answer but aren't marked as gold passages, which is consistent with the findings from related studies~\cite{arabzadeh2021shallow}. 
Overall, we find that \textsc{ELBoLoss} improves relevance of retrieved passages over \textsc{MarginalizedLoss} for two qualitatively different tasks, with larger gains for the one-to-many generation task.

\subsection{Groundedness Evaluation}
\label{sec:groundedness_evaluation}

\begin{table}
  \caption{Groundedness evaluation: Our method \textsc{ELBoLoss} ($\alpha=0.25$) shows more overlap between generated output and the retrieved passage than \textsc{MarginalizedLoss} and for the Wizard of Wikipedia dataset the gap increases as we consider the maximum over top-$5$ passages. (Novel-F1: discounts commonly occurring words and context words ($x$))}
  \label{table:groundedness_evaluation}
  \centering
  \begin{tabular}{ll|rr|rr}
    \toprule
    & & \multicolumn{2}{c|}{Top-1} & \multicolumn{2}{c}{Max.\ of Top-5} \\
    Dataset & Method                                       &  F1            & Nov-F1        & F1              & Nov-F1  \\
    \midrule
     \multirow{2}{*}{WoW}   & Marg. Gen.                   & 18.63          & 17.46        & 26.19          & 25.39  \\ %
                            & ELBo Gen. ($\alpha=0.25$)   & \textbf{21.34} & \textbf{20.78} &  \textbf{34.16} & \textbf{34.24} \\ %
        \midrule
     \multirow{2}{*}{MSM}   & Marg. Gen.              & 33.12 & 25.39    & 45.76 & 39.45 \\ %
                                & ELBo Gen.($\alpha=0.5$) & \textbf{34.52}  & \textbf{26.49} &  \textbf{46.91} & \textbf{40.47}
 \\ %
    \bottomrule
  \end{tabular}
\end{table}

We now examine \textbf{RQ$_2$}, studying the degree to which the generator relies on the retrieved passages for producing its output. To quantify this \textit{groundedness}, we compute F1-overlap between a \textit{retrieved passage} (not necessarily the gold passage) and the produced text when generation is conditioned on that passage.

As an analogue of Success@$k$, we propose \textit{Max.\ F1@k}, the largest F1-overlap exhibited by any generated output with the corresponding retrieved passage fed to the generator. We also propose Novel-F1, a new metric that discounts words the occur frequently and words that already appear in the context $x$, since otherwise these tokens dominate raw F1 in practice (up to 80\%, see Appendix~\ref{sec:novel-f1}). We restrict this to the novel words, since it is easy for the generator to ``copy'' common words without that indicating grounding in the content of the passage.

Our results are shown in Table~\ref{table:groundedness_evaluation} and Figure~\ref{fig:relevance_groundedness_comparison}. 
For the Wizard of Wikipedia dataset, we observe that our \textsc{ELBoLoss} generator  outperforms \textsc{MarginalizedLoss} by 2.7 F1 (14.5\% r.i.) and 3.3 Novel-F1  (19\% r.i.) when given the top retrieved passage. 
In Figure~\ref{fig:relevance_groundedness_comparison} (right), we can see that beyond the top passage, \textsc{MarginalizedLoss} generator's overlap decays rapidly whereas \textsc{ELBoLoss} ($\alpha=0.25$) generator's overlap declines gradually. 
This shows that the \textsc{ELBoLoss} generator stays grounded beyond just the top passage, a desirable property in one-to-many generation systems.
The above plot could hide the scenario where \textsc{MarginalizedLoss} is grounded in just one out of the top-$k$ passages (say the gold passage), reducing the average overlap while still being grounded. 
Therefore we consider Max.\ F1@$k$ and Max Novel-F1@$k$ and observe that \textsc{ELBoLoss} generator outperforms \textsc{MarginalizedLoss} by 8 F1@5 and 8.8 Novel-F1@5, which shows that even when allowed to pick the most suitable passage, \textsc{MarginalizedLoss} heavily underperforms. 
For the MS MARCO NLGen dataset, we observe smaller but consistent gains in groundedness (1--2 F1, Novel-F1) with \textsc{MarginalizedLoss} compared to \textsc{ELBoLoss}. This is expected because for a QA task $\Pret$ is a good proxy for label-relevance and \textsc{ELBoLoss} gains little by optimizing the expectation of the log-likelihood over $\Q$. Overall, we see that \textsc{ELBoLoss} is more grounded than \textsc{MarginalizedLoss}, with a bigger margin for one-to-many tasks.

\subsection{End-to-end Evaluation}
\label{sec:generation_quality_evaluation}
To evaluate the end-to-end quality of our systems, we calculate F1 and Novel-F1 (defined in Section~\ref{sec:groundedness_evaluation}) of the decoded output with the \textit{human-written gold output}. To allow for the possibility of the generator using any part of the long passages (150 words) for the Wizard Of Wikipedia task, we use Knowledge-F1 (defined by \cite{shuster2021retrieval}): F1 between the sampled generation and the \textit{gold passage}.
Since it is reasonable to expect the gold passage to be in top-$k$ for $k$>1 for one-to-many tasks (as in Section~\ref{sec:groundedness_evaluation}), we also compute the Max.\ over top-$k$ retrieved passages.  %

The results are summarized in Table~\ref{table:e2e_evaluation}. For Wizard of Wikipedia, we see that using only the top retrieved passage \textsc{ELBoLoss} only slightly improves performance over \textsc{MarginalizedLoss}. This is expected in the one-to-many setting: the label-relevant passage is an arbitrary choice from amongst the context-relevant passages, and the $6.7\%$ relative improvement in Novel-F1@$1$ is consistent with the relatively small improvements in relevance and grounding for the rank $1$ passage.
We see larger improvements for \textsc{ELBoLoss} with Max.\ overlap over the top-$5$ passages, namely 1~F1, 2~Novel-F1 ($\sim15\%$ r.i.), and 2~K-F1 ($>10\%$ r.i.). 
This shows the retriever and generator improvements for $k>1$ contribute to a better end-to-end system, with one of the generated outputs having a higher overlap with the target output because it was better grounded in a label-relevant passage in the top-$k$ retrieved passages. For MS Marco NLGen, we see a small but consistent increase due to \textsc{ELBoLoss} over \textsc{MarginalizedLoss}: 1.5 F1 and 2 Novel-F1 across passage depths. 

We also submit models trained using \textsc{ELBoLoss} and \textsc{MarginalizedLoss} on Wizard of Wikipedia to the KILT leaderboard. The results are reported in Table~\ref{table:leaderboard_evaluation}. As shown, \textsc{ELBoLoss} consistently outperforms the baseline \textsc{MarginalizedLoss} across all metrics. The table also reports Recall@5, which evaluates retrieval at a coarser granularity, namely at the \textit{full Wikipedia page} level, though so far we have investigated it directly at the passage level. Consistent with the results in Table~\ref{table:relevance_evaluation}, our method also outperforms \textsc{MarginalizedLoss} in retrieval metrics. In fact, our \textsc{ELBoLoss} model achieves state-of-the-art performance across all the generation metrics (F1, ROUGE-L, KILT-F1, KILT-ROUGE-L) on the leaderboard, though it is not the strongest on Recall@5.\footnote{Earlier results on the KILT leaderboard for Wizard of Wikipedia should be interpreted with caution, as the KILT authors recently updates the train/dev splits due to anomalies in the preprocessing script. We have used the updated version for our model and baseline.}

To conclude, we have evaluated the \textsc{ELBoLoss} and \textsc{MarginalizedLoss} using a one-to-one free-form QA dataset and a one-to-many dataset of informative conversations. Our results show that our method \textbf{ELBoLoss} trains a better retriever, a more grounded generator and improves end-to-end performance, especially in the one-to-many setting. 

\begin{table}
  \caption{Wizard of Wikipedia KILT leaderboard evaluation: \textsc{ELBoLoss} achieves SoTA on generation metrics (F1, ROUGE-L, KILT-F1, KILT-ROUGE-L indicated with $\dagger$) as of Oct 2021 and improves relevance over \textsc{MarginalizedLoss}} 
  \label{table:leaderboard_evaluation}
  \centering
  \begin{tabular}{lrrrrrr}
    \toprule
        & R-Prec & Recall@5 & F1 & ROUGE-L & KILT-F1 & KILT-ROUGE-L \\
    \midrule
     Marg.       & 53.94  & 68.12 & 18.11 & 16.21   & 11.78   & 10.47 \\
     ELBo      & \textbf{56.08} & \textbf{74.26} & \textbf{19.19}\rlap{$^{\dagger}$} & \textbf{17.06}\rlap{$^{\dagger}$} & \textbf{13.39}\rlap{$^{\dagger}$} & \textbf{11.92}\rlap{$^{\dagger}$} \\

    \bottomrule
  \end{tabular}
\end{table}
\begin{table}
  \caption{End-to-end evaluation: Our method \textsc{ELBoLoss} improves over \textsc{MarginalizedLoss} when considering Max.\ overlap of generated output with target output over top-$5$ passages for the Wizard of Wikipedia dataset and also for top-$1$ with MS Marco NLGen dataset. (Novel-F1: discounts commonly occurring words and context words ($x$), Knowledge-F1: overlap of generated output with gold passage)}
  \label{table:e2e_evaluation}
  \centering
  \begin{tabular}{ll|rrr|rrr}
    \toprule
    & & \multicolumn{3}{c|}{Top-1}  & \multicolumn{3}{c}{Max.\ of Top-5}  \\
    Dataset & Method                                      &  F1    & N-F1 & K-F1  & F1    & N-F1 & K-F1 \\
    \midrule
      \multirow{2}{*}{WoW}      & Marg.              & 18.79          & 10.45           & 12.61      & 26.52       & 16.42 &    12.45\\
         & ELBo                     & \textbf{18.86} & \textbf{11.12}  & \textbf{13.08} & \textbf{27.56} & \textbf{18.67} & \textbf{14.62} \\
        \midrule
    \multirow{2}{*}{MSM }  & Marg.                & 60.18           & 37.19          & --     & 72.22            & 56.06
           & -- \\
     & ELBo                          & \textbf{61.46}  & \textbf{39.65} & --     &  \textbf{73.18} & \textbf{58.19}  & -- \\
    \bottomrule
  \end{tabular}
\end{table}

\section{Discussion}

\paragraph{Hallucination, grounding and correctness}
\cite{shuster2021retrieval} show that providing retrieved passages to a generator has been shown to reduce hallucination. Our work increases grounding in the retrieved passage, promising to further reduce hallucination. Even though the generator is now more likely to use content from the provided passage (rather than hallucinating from parametric memory), that does not guarantee correctness.%
This is not captured perfectly by our token-level overlap metrics that evaluate grounding. There is scope for future work to address this gap with better generator training methods that not only produce grounded but also correct outputs. 

\paragraph{Passage selection for controllable generation} 
Our results show that the \textsc{Hindsight} generator stays grounded beyond the top retrieved passage. 
As a consequence, we can exert considerable control over the generated content by controlling the passage provided to the generator, potentially complementing controllable language generation using special tokens or attributes \citep{PPLM}. 
This can have significant impact when such systems are deployed in real-life settings (e.g., in open-domain socialbots) where external business logic with concrete objectives can help select an appropriate passage from the top-$k$ retrieved passages and control the generator's output.

\paragraph{Modulating trust}
For tasks like QA, we want a ``conservative'' generator: it should abstain from using a passage that doesn't contain the answer. For more open-ended tasks like informative conversations, we would like the generator make use of diverse passages. In the first case, we want the generator to ``trust'' the retrieved passages less and in the second case, more . 
In our work, we show that by changing the distribution of the passage from $P_\eta(.|x)$ to $Q(.|x,y)$, the generator increasingly trusts the retrieved passages (Figure~\ref{fig:relevance_groundedness_comparison}).
Based on the nature of the task, system designers can use the  $\alpha$-mixture to modulate the degree of trust placed by the generator in the retrieved passages.

\paragraph{Comparison with Fusion-in-Decoder}
\cite{FiD} provide multiple passages to the generator simultaneously and in \cite{FiDKD} they use the decoder's attention weights over each passage for relevance supervision. While this is a valid mechanism, perhaps conditioning on individual passages like we do is more precise for relevance supervision. Indeed, recent work \citep{EMDR} illustrates by using Fusion-in-Decoder during inference but foregoing the decoder's attention weights and using an equivalent version of the \textsc{MarginalizedLoss} for training the retriever. Furthermore, Fusion-in-Decoder is uniquely useful for QA style tasks, where it has to select the correct answer from many passages. But for one-to-many tasks, our system can condition on each passage separately and generate diverse outputs whereas with Fusion-in-Decoder that cannot happen. 

In this paper, we propose \textsc{Hindsight}, a system that introduces a guide-retriever to improve supervision for both the retriever and the generator for retrieval-augmented, open-ended generation. During training, the guide retriever is allowed to use the target output of each example---alongside the input context---in order to find relevant passages, improving the coverage of useful passages during training and, in turn, leading to better retrieval and more grounded generation. The resulting system achieves considerable empirical improvements over existing work, improving retrieval quality by up to 23\%, grounding by up to 19\%, and end-to-end output quality by up to 6.4\%.

\bibliography{hindsight}
\bibliographystyle{iclr2022_conference}

\clearpage
\appendix

\section{Appendix}
\subsection{Examples of retrieved passages}
\label{sec:retrieved_passage_examples}
\subsubsection{Conversation 1: Italian cuisine}

\begin{description}\setlength{\itemsep}{0pt} 
\item[Other] Ooh I like that! Stick some nice spicy arrabbiata sauce with it, ahhhh! Have you ever had bucatini before?
\item[Self] Oh yeah! I love that spicy garlic and tomato sauce. No I have not had bucatini. Is that a type of cheese?
\item[Other] Now you're speakin' my language. No no, it's a style of noodle, like a really long straw. Bucatini amatraciana is insanely good.
\end{description}

\begin{table}
    \centering
        \begin{tabularx}{\linewidth}{lX}
        \toprule
        rank &    text                                                                                                                                                                                                      \\
        \midrule
        1.0  &  Bucatini > Abstract | Bucatini , also known as perciatelli , is a thick spaghetti-like pasta with a hole running through the center. The name comes from , meaning "hole", while "bucato" or its Nea... \\
        2.0  &  Pasta con le sarde > Ingredients. | The principal ingredients are olive oil, onions, pasta and a finely chopped mixture of sardines and anchovy. Various types of pasta are used for the dish, but b... \\
        3.0  &  Bucatini > Preparation. | Standard pasta machines will roll out sheets of flat pasta which are then cut into ribbons to make flat, ribbon-style pasta like fettuccine, tagliatelle, or pappardelle. ... \\
        4.0  &  Bocconcini > Abstract | This cheese is described by its Italian name, which means "small mouthfuls". It is made in the "pasta filata" manner by dipping curds into hot whey, and kneading, pulling, ... \\
        5.0  &  Carbonara > Abstract | Carbonara () is an Italian pasta dish from Rome made with egg, hard cheese, guanciale (or pancetta), and black pepper. The dish arrived at its modern form, with its current ... \\
        \bottomrule
        \end{tabularx}
    \caption{Passages retrieved by \textsc{ELBoLoss} retriever while talking about Italian Cuisine. They include passages about various ingredients (rank=2), cheeses (rank=4), dishes (rank=5) alongside more information about Bucatini Pasta (rank=1,3).}    
    \label{tab:my_label}
\end{table}

\begin{table}
    \centering
        \begin{tabularx}{\linewidth}{lX}
        \toprule
        rank & text \\
        \midrule
        1.0  &  Bucatini > Abstract | Bucatini , also known as perciatelli , is a thick spaghetti-like pasta with a hole running through the center. The name comes from , meaning "hole", while "bucato" or its Nea... \\
        2.0  &  Bucatini > Preparation. | Standard pasta machines will roll out sheets of flat pasta which are then cut into ribbons to make flat, ribbon-style pasta like fettuccine, tagliatelle, or pappardelle. ... \\
        3.0  &  Rotini > Abstract | Rotini is a type of helix- or corkscrew-shaped pasta. The name comes from a 17th-century Italian word meaning "small wheels". Rotini is related to fusilli, but has a tighter he... \\
        4.0  &  Vermicelli > History.:The Americas. | The "fideo" is a type of noodle, produced in Europe ever since the Roman times, best known as "fideus" or "fidelis", and then spread to Mexican and Latin Amer... \\
        5.0  &  Rollatini > Abstract | Rollatini (sometimes also spelled rolatini or rolletini) is an Italian-style dish (called "rollatini di melanzane" in faux Italian) that is usually made with thin slices of ... \\
        \bottomrule
        \end{tabularx}
    \caption{Passages retrieved by \textsc{MarginalizedLoss} retriever while talking about Italian Cuisine. All passages talk about Pastas.}
    \label{tab:MargPasta}
\end{table}
\clearpage 
\subsubsection{Conversation 2: Rock and Roll}

\begin{description}\setlength{\itemsep}{0pt} 
\item[Self] Do you mean  Elvis Aaron Presley, the American singer and actor? 
\item[Other] That's the one. I think his nickname was the king of rock 'n roll.
\item[Self] I had just heard of him being "The King".  There probably would not have been a Sun Records if not for Elvis and Sam Phillips.
\item[Other] He was revolutionary for his time. Many older people thought he was straight from the devil.
\end{description}
\begin{table}[h]
    \centering
        \begin{tabularx}{\linewidth}{lX}
        \toprule
        rank &   text        \\
        \midrule
        1.0  &  Sam Phillips > Abstract | Samuel Cornelius Phillips (January 5, 1923 – July 30, 2003) was an American record producer who played an important role in the development of rock and roll during the 19... \\
        2.0  &  Cultural impact of Elvis Presley > Abstract | Since the beginning of his career, Elvis Presley has had an extensive cultural impact. According to "Rolling Stone", "it was Elvis who made rock 'n' r... \\
        3.0  &  Freddie King > Abstract | Freddie King (September 3, 1934 – December 28, 1976) was an American blues guitarist and singer. He recorded several hits for Federal Records in the early 1960s. His soul... \\
        4.0  &  Elvis Presley > Abstract | With a series of successful network television appearances and chart-topping records, he became the leading figure of the newly popular sound of rock and roll. His energ... \\
        5.0  &  Elvis Presley > Abstract | Elvis Aaron Presley (January 8, 1935 – August 16, 1977), also known mononymously as Elvis, was an American singer, musician, and actor. Regarded as one of the most signi... \\
        \bottomrule
        \end{tabularx}
    \caption{Passages retrieved by \textsc{ELBoLoss} while talking about Rock and Roll. Relevant passages about cultural impact of Elvis Presley (rank=2) and details about his career (rank=4) alongside introductary paragraphs of other musicians}
    \label{tab:elbo_rock}
\end{table}

\begin{table}[h]
    \centering

\begin{tabularx}{\linewidth}{lX}
\toprule
rank &    text                                                                                                                                                                                                      \\
\midrule
1.0  &  Elvis Presley > Abstract | Elvis Aaron Presley (January 8, 1935 – August 16, 1977), also known mononymously as Elvis, was an American singer, musician, and actor. Regarded as one of the most signi... \\
2.0  &  Sam Phillips > Abstract | Samuel Cornelius Phillips (January 5, 1923 – July 30, 2003) was an American record producer who played an important role in the development of rock and roll during the 19... \\
3.0  &  Johnny Otis > Abstract | Johnny Otis (born Ioannis Alexandres Veliotes; December 28, 1921 – January 17, 2012) was an American singer, musician, composer, arranger, bandleader, talent scout, disc j... \\
4.0  &  Carl Perkins > Abstract | Called "the King of Rockabilly", he was inducted into the Rock and Roll Hall of Fame, the Rockabilly Hall of Fame, the Memphis Music Hall of Fame, and the Nashville Songw... \\
5.0  &  Chubby Checker > Abstract | Chubby Checker (born Ernest Evans; October 3, 1941) is an American rock 'n roll singer and dancer. He is widely known for popularising many dance styles including the t... \\
\bottomrule
\end{tabularx}

    \caption{Passages retrieved by \textsc{MarginalizedLoss} while talking about Rock and Roll. All passages are the introductory paragraphs from various related artists}
    \label{tab:my_label_two}
\end{table}
\subsection{Novel-F1}
\label{sec:novel-f1}
\paragraph{Rationale} We conducted a small experiment with the generated output on Wizard of Wikipedia dataset using top-8 retrieved passages. We removed the gold passage and computed overlap of the generated output with the target output. We consistently found (across models and passage ranks) the F1 overlap to be close to 15. This meant that by conditioning on arbitrary passages the generator (likely by ignoring them altogether) is able to achieve around 80\% of the F1-overlap of the best performing models ($\sim19$ F1). This can be a confounding factor for selecting models based on high F1 overlap. A model that simply copies content from the input $x$ can achieve high F1-overlap but fail to using the retrieved passage to generate the output. Removing commonly occurring words reduces it to 8 F1, but removing words from input context reduces it further down to 4 F1. Thus we find Novel-F1 to be the cleanest measure of overlap as it discounts two confounding factors and only looks at ``Novel'' words, words that are rare and were not in the input text $x$.    

We construct the list of common words based on their frequency in the training corpus. We sort words by frequency and take the most frequent words that contribute: 50\% of the probability mass toward Wizard of Wikipedia utterances (amounting to 121 words) following \cite{shuster2021retrieval}. However, we found that using the same heuristic for MS-MARCO NLGen answers included numbers and rarer tokens that could potentially be in the answer span. So we instead use only 33\% of the probability mass (amounting to 55 words). We also ran evaluation using 50\% of the probability mass but found the trends to be consistent. 

\paragraph{MS Marco NLGen list of common words} (sorted by frequency) \\
is, of, in, to, and, for, or, are, that, on, from, as, by, you, with, it, county, can, at, per, was, your, average, cost, be, between, which, used, one, united, states, there, years, located, name, not, new, have, takes, number, has, means, days, when, blood, system, year, should, no, most, first, hours, up, minutes, 1

\paragraph{Wizard of Wikipedia list of common words} (sorted by frequency) \\
is, of, in, to, and, for, or, are, that, on, from, as, by, you, with, it, county, can, at, per, was, your, average, cost, be, between, which, used, one, united, states, there, years, located, name, not, new, have, takes, number, has, means, days, when, blood, system, year, should, no, most, first, hours, up, minutes, 1
i, and, of, in, is, to, it, that, are, you, they, have, was, but, for, as, its, like, with, on, so, be, or, not, yes, do, can, from, there, by, well, also, one, my, know, has, some, he, their, love, most, people, think, really, all, about, just, too, them, im, which, sure, more, been, at, would, many, were, good, very, dont, when, thats, no, yeah, what, other, great, if, because, used, actually, first, since, lot, me, even, your, how, we, time, different, world, use, get, called, only, out, much, over, had, though, music, around, popular, his, am, made, than, such, back, up, us, make, usually, who, favorite, new, food, oh, long, she, now, did, pretty, any, where, years, this, way, go

\section{Intuition behind improvements due to ELBoLoss}
\begin{figure}[ht]
  \centering
    \subfloat{\includegraphics[width=2.5in]{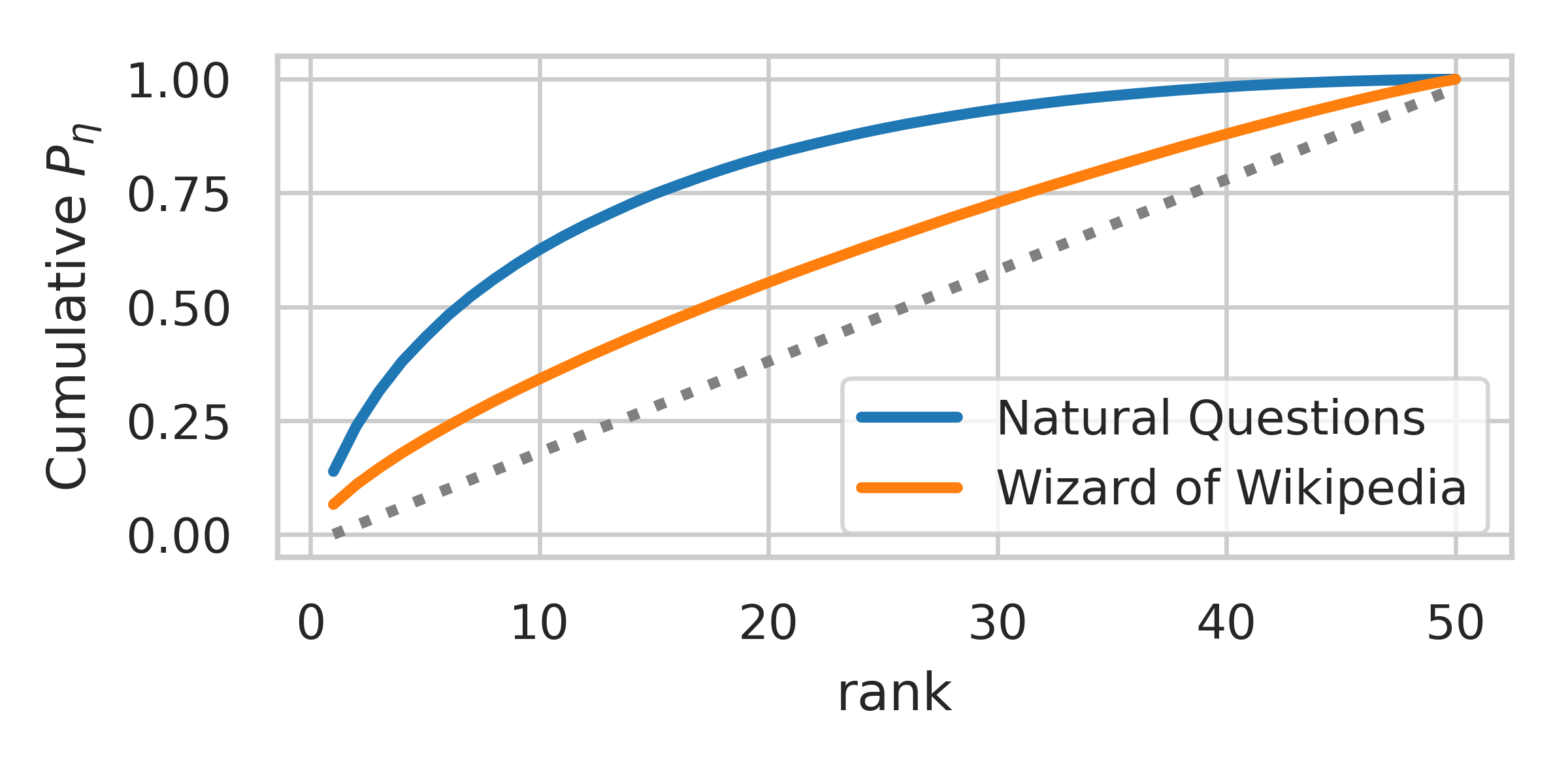}}
    \qquad
    \subfloat{\includegraphics[width=2.5in]{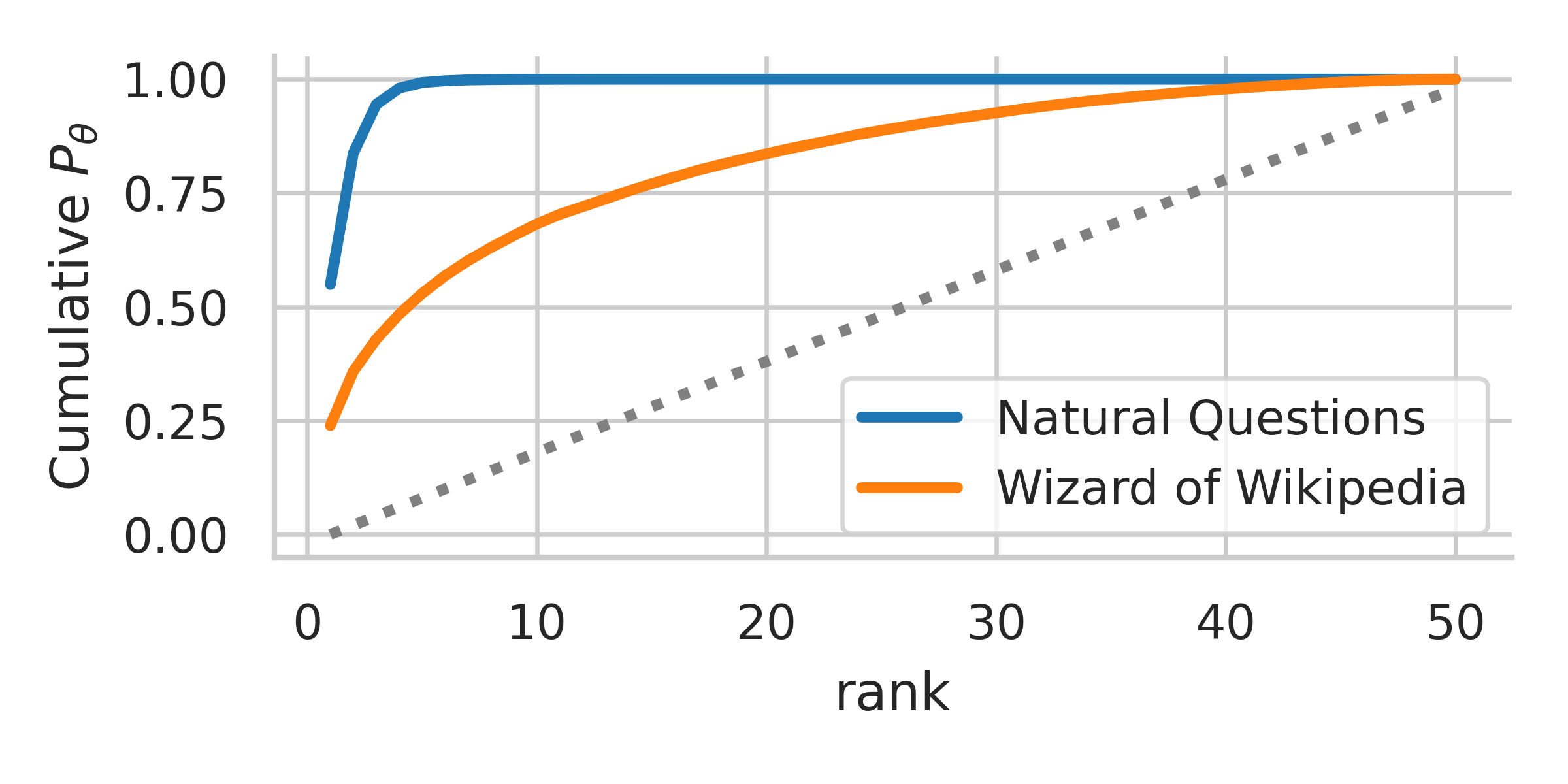}}

    \caption[]{With \textsc{MarginalizedLoss}, the generator $\Pgen$ learns a sharp distribution for Natural Questions (NQ) dataset (\textbf{right}) but learns a flatter distribution for a one-to-many open-ended generation task using the Wizard of Wikipedia dataset (WoW). The flatter distribution in the case of WoW Generator shows that it has not learned label-relevance as well. Consequently, for WoW we see a weaker retriever (\textbf{left}) that has a flatter distribution than NQ. (\textbf{Left}) Cumulative probability $\Pret$ w.r.t.\ rank for passages. 
    (\textbf{Right}) Assuming a uniform prior $P(z|x)$, the cumulative probability $\Pgen$ w.r.t.\ rank for passages, plotted as $P(z|x,y) \propto P(y|x,z) P(z|x)$.
    The gray dotted line shows a hypothetical model that assigns equal probabilities to all passages.} 
    \label{fig:attention_collapse}
\end{figure}

To understand the intuition behind suboptimality of \textsc{MarginalizedLoss} for open-ended generation tasks consider the following: 
We would want a good retriever to assign similar but high probabilities to all context-relevant passages because they are similarly relevant but a good generator to only assign high probabilities when using label-relevant passages because only label-relevant passages are pertinent to the target output. 
But the training signal to a model (partial derivative w.r.t the model and a passage) is modulated by the probability of the other model: 
\begin{align*}
\frac{\partial P(y|x)}{\partial P_\eta(z_i|x)} &= P_\theta(y|x,z_i) & \frac{\partial P(y|x)}{\partial P_\theta(y|x, z_i)} &= P_\eta(z_i|x) 
\end{align*}
Since context-relevant passages have similar $P(z_i|x)$ the gradient encourages the generator to assign equal probabilities to the target output using all context-relevant passages. 
We see this issue play out empirically when using \textsc{MarginalizedLoss} for two different tasks: Open-Domain QA (Natural Questions by \cite{NQ}) and informative dialogue (Wizard of Wikipedia by \cite{WoW}) (Figure~\ref{fig:attention_collapse}). 
We see that on the Natural Questions dataset, where there is typically one correct answer, the generator produces distribution with a sharp peak that can potentially serve as an accurate proxy for label-relevance and in turn train a good retriever. 
But on the Wizard of Wikipedia dataset, the generator produces a flatter distribution which is a bad proxy for label-relevance. This provides weaker supervision for the retriever which learns a flatter probability distribution as well and is less able to differentiate context-relevant from irrelevant passages. 

\begin{figure}
  \centering
    \subfloat{{\includegraphics[width=2.5in]{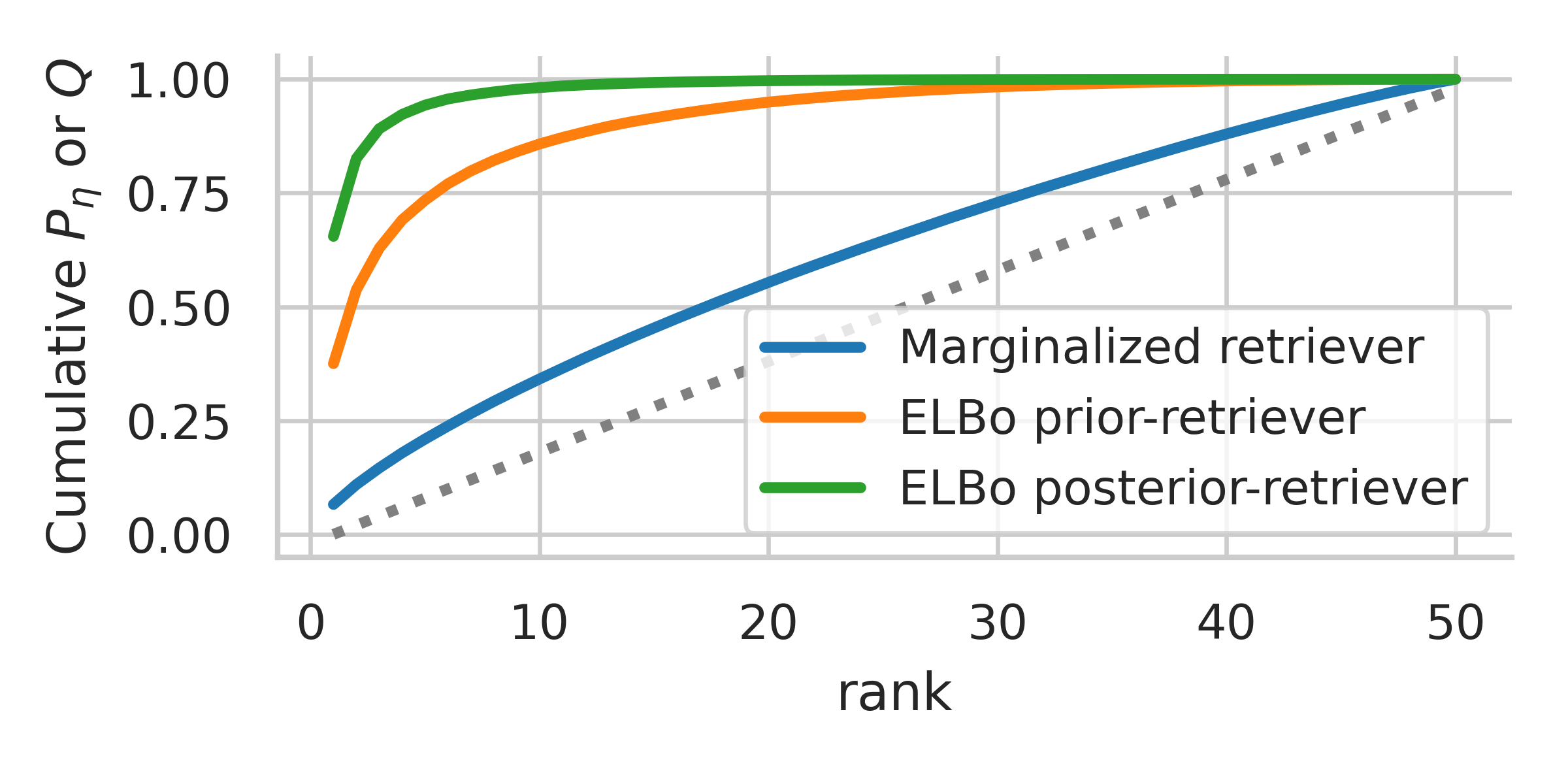}}}
    \qquad
    \subfloat{{\includegraphics[width=2.5in]{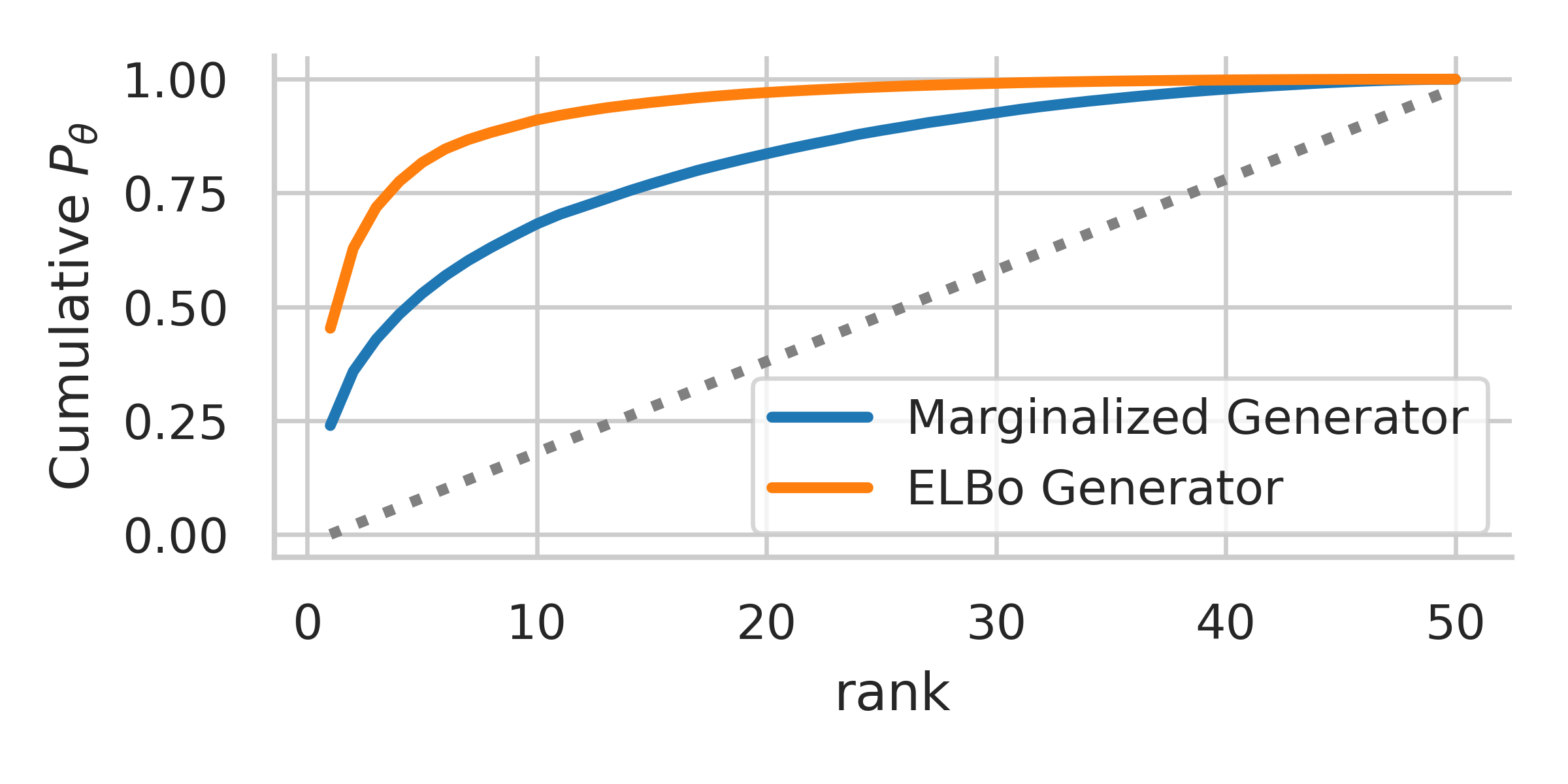}}}
    \caption[]{Analogous plots to Figure~\ref{fig:attention_collapse} but with \textsc{ELBoLoss} on the one-to-many Wizard of Wikipedia (WoW) dataset. Training with \textsc{ELBoLoss} produces a sharp distribution for $\Q$ and subsequently sharper $\Pret$ and $\Pgen$ than \textsc{MarginalizedLoss}. }
    
    \label{fig:improved_attention}
\end{figure}

We see in Figure~\ref{fig:improved_attention} that for the Wizard of Wikipedia dataset with \textsc{ELBoLoss} we obtain a sharp distribution for $\Q$ (nearly as good as $\Pgen$ on NQ from Figure~\ref{fig:attention_collapse}) and that the $\Pret$ and $\Pgen$ are now sharper than \textsc{MarginalizedLoss}.
While a sharper distribution does not imply a better retriever and generator (they may still assign high probability to the wrong passage), a flatter distribution limits their potential. 
As we will see in Section~\ref{sec:exp_eval}, \textsc{ELBoLoss} indeed utilizes the potential and trains a better retriever and more a grounded generator.

\end{document}